\documentclass{article} 
\usepackage{iclr2027_conference,times}

\usepackage{amsmath,amsfonts,bm}

\def\eqref#1{equation~\ref{#1}}

\def\1{\bm{1}}

\DeclareMathAlphabet{\mathsfit}{\encodingdefault}{\sfdefault}{m}{sl}
\SetMathAlphabet{\mathsfit}{bold}{\encodingdefault}{\sfdefault}{bx}{n}

\usepackage[hidelinks]{hyperref}
\usepackage{url}

\usepackage{amsmath}
\usepackage{graphicx}
\usepackage{xcolor}
\usepackage{booktabs}
\usepackage{multirow}
\usepackage{arydshln}
\usepackage{pifont}
\usepackage[most]{tcolorbox}
\usepackage{cleveref}

\newtcolorbox{promptbox}[1]{
  enhanced,
  breakable,
  colback=gray!4,
  colframe=black!35,
  colbacktitle=gray!18,
  coltitle=black,
  boxrule=0.4pt,
  arc=1pt,
  left=4pt,
  right=4pt,
  top=4pt,
  bottom=4pt,
  title={#1},
  fonttitle=\bfseries\small,
  fontupper=\small\itshape,
  before upper={\setlength{\parindent}{0pt}}
}
\newtcolorbox{supervisionbox}[1]{
  enhanced,
  breakable,
  colback=blue!2,
  colframe=blue!35!black,
  colbacktitle=blue!10,
  coltitle=black,
  boxrule=0.4pt,
  arc=1pt,
  left=4pt,
  right=4pt,
  top=4pt,
  bottom=4pt,
  title={#1},
  fonttitle=\bfseries\small,
  fontupper=\small,
  before upper={\setlength{\parindent}{0pt}}
}
\newcommand{\promptslot}[1]{%
  \begingroup
  \setlength{\fboxsep}{1pt}%
  \setlength{\fboxrule}{0.3pt}%
  \fbox{\normalfont\small #1}%
  \endgroup
}

\title{Advancing Video-Text Pretraining with Multi-View Captions}
\author{
\centerline{\textbf{Fida M. Thoker}$^{1}$\thanks{These authors contributed equally.}\hspace{8pt}
  \textbf{Renaud Vandeghen}$^{2}$\footnotemark[1]\hspace{8pt}
  \textbf{Karen Sanchez}$^{1}$}
\vspace{1pt}\and\vspace{2pt}\centerline{\textbf{Marc Van Droogenbroeck}$^{2}$\hspace{8pt}
  \textbf{Bernard Ghanem}$^{1}$}
\and\centerline{$^{1}$ King Abdullah University of Science and Technology (KAUST)}
\and\centerline{$^{2}$ University of Liège}
}

\iclrfinalcopy 
\begin{document}

\maketitle

\begin{abstract}
Video-text pretraining has achieved remarkable progress through the scaling of models and datasets, yet the quality of language supervision remains underexplored. Existing web-scale datasets often provide only a single sparse caption per video that fails to capture rich spatiotemporal semantics, while directly using captioning models can generate noisy descriptions. We propose a large-scale multimodal large language model-based supervision generation framework that improves supervision diversity, fidelity, and semantic coverage. Starting from 10 million videos, our approach generates \textbf{m}ulti-\textbf{v}iew \textbf{c}aptions (MVC) through complementary summary and detailed captions, reasoning-based refinement, and semantic positive caption generation.
To effectively exploit supervision at different granularities, we further introduce a granularity-aware text representation with separate \texttt{CLS} tokens for summary and detailed views.
We pretrain video-text models using the resulting supervision corpus and evaluate them across standard, fine-grained and detailed text-to-video retrieval benchmarks. Our approach consistently improves both zero-shot and fine-tuned performance while using smaller pretraining corpora than existing methods, demonstrating the importance of rich and complementary textual supervision for video-text pretraining. Project page: \url{https://rvandeghen.github.io/mvc/}
\end{abstract}

\vspace{-0.3cm}
\section{Introduction}
\label{sec:introduction}

Video-text pretraining has emerged as a fundamental paradigm for learning transferable video representations~\citep{xu2021videoclip,wang2022internvideo,bain2021frozen,wang2023all,Li2023Unmasked,lei2023revealing,Wang2024InternVid,Wang2024InternVideo2,doughty2024locomotion}. By aligning videos with natural language at scale, these approaches learn semantic representations that generalize across diverse downstream tasks, including retrieval, captioning, video question answering, etc. Inspired by the success of image-text pretraining~\citep{Radford2021Learning,jia2021scaling,li2021align,Li2022BLIP}, recent efforts have scaled both video-text corpora and model capacity~\citep{Wang2024InternVid,Wang2024InternVideo2,chen2024panda}, establishing scaling as a dominant paradigm for advancing video-text learning.

Despite this progress, most works primarily improve performance through larger datasets, stronger objectives, and increased model capacity, while the quality of language supervision itself remains comparatively underexplored. Existing video-text datasets are largely collected from web~\citep{bain2021frozen,miech2019howto100m,Wang2024InternVid} and rely on captions that are often short, sparse, weakly aligned, or noisy. As a result, rich video content is frequently compressed into a single high-level description that discards important visual cues. For example, a caption such as ``a person playing football'' captures the dominant activity while overlooking temporal progression, object interactions, and contextual details crucial for learning transferable representations. Since videos naturally contain multiple objects, actions, interactions, and evolving events, reducing them to a single description encourages models to learn coarse semantic correspondences while overlooking fine-grained dynamics and complementary relationships. This reveals a fundamental supervision bottleneck in current video-text pretraining: the limitation arises not only from insufficient information, but also from representing rich visual content through a single textual view. This naturally raises an important question: \emph{can rich video content be effectively learned from a single textual view?} A video may admit multiple valid semantic interpretations, yet current supervision compresses these complementary signals into a single caption.

A natural solution is to leverage multimodal large language models (MLLMs) to generate richer supervision automatically. Recent studies have demonstrated the potential of large-scale recaptioning pipelines~\citep{chen2024panda,Zheng2024DreamLIP,chen2024sharegpt4video,chen2024sharegpt4v}. However, generating longer captions alone may not necessarily lead to better supervision, as such descriptions may emphasize irrelevant details, introduce incorrect information, or drift from the dominant visual content. Effective video-text supervision, therefore, requires signals that are not only richer, but also \emph{diverse}, \emph{visually faithful}, and \emph{semantically comprehensive}.

In this work, we revisit video-text pretraining from the perspective of supervision quality and propose a fully automated MLLM-based framework for generating richer and more reliable textual supervision at scale. Given an input video, we generate summary and detailed descriptions using multiple instruction-tuned MLLMs, refine them through reasoning-based visual verification, and construct additional semantic positive captions capturing complementary perspectives. To effectively exploit supervision at different levels of granularity, we further introduce a granularity-aware text representation with separate \texttt{CLS} tokens for summary and detailed views, allowing them to be independently aligned with the video while learning complementary representations. Using this framework, we re-caption approximately 10 million videos to construct a large-scale video-text pretraining corpus. Extensive experiments across standard, fine-grained, and detailed video-text retrieval benchmarks demonstrate that \textbf{m}ulti-\textbf{v}iew \textbf{c}aptions (MVC) substantially improve video-text pretraining and provide a data-efficient alternative to simply scaling the number of video-text pairs.

\vspace{-0.2cm}
\section{Related Work}
\label{sec:related}

\vspace{-0.1cm}
\paragraph{Video-Text Pretraining.}
Video-text pretraining has evolved from early approaches that learn aligned video-text representations through contrastive learning and cross-modal matching objectives~\citep{bain2021frozen,lei2021less,xu2021videoclip}. Building on these, subsequent works improved representation quality through larger-scale video-text datasets~\citep{lei2023revealing}, stronger visual encoders and multimodal architectures~\citep{wang2022omnivl,wang2022internvideo,liu2022umt,Li2023Unmasked}, scaling strategies~\citep{Wang2024InternVideo2}, and more effective masking~\citep{zhuang2026cluster,wu2025video}. These developments have substantially advanced transfer performance across retrieval and video understanding tasks.

More recent efforts have further shifted toward large-scale video-text foundation models trained on millions or even billions of video-text pairs~\citep{Wang2024InternVid,chen2024panda,Wang2024InternVideo2}. While these approaches largely improve performance through model and data scaling, relatively few works investigate the role of supervision design itself. Our work differs from existing approaches by focusing on the supervision source and exploring whether richer, multi-view textual supervision can provide stronger learning signals for video-text pretraining.

\vspace{-0.2cm}
\paragraph{MLLM-Based Caption Generation.}
Several works have explored synthetic supervision to improve language quality by generating or rewriting captions using generative models and large language models, aiming to reduce noisy supervision and create more informative and grounded textual descriptions~\citep{Wang2024InternVid,Yuan2025Tarsier2-arxiv,islam2024video,fan2023improving}.

In the image-text domain, ShareGPT4V~\citep{chen2024sharegpt4v} trains a captioning model on a curated set of image-caption pairs and uses it to generate large-scale detailed descriptions. SynthCLIP~\citep{hammoud2024synthclip} creates synthetic image-text pairs with text-to-image generation while LaCLIP~\citep{fan2023improving} rewrites existing captions as a form of textual augmentation. DreamLIP~\citep{Zheng2024DreamLIP} further demonstrates that recaptioning images with off-the-shelf MLLMs can significantly improve downstream performance by generating richer textual supervision.
For the video domain, ShareGPT4Video~\citep{chen2024sharegpt4video} extends large-scale recaptioning to videos by first collecting and annotating approximately 40K high-quality video-caption pairs, which are then used to train a video captioner for generating 4.8M detailed video-text pairs. InternVid~\citep{Wang2024InternVid} constructs video annotations by generating frame-level captions using BLIP-2~\citep{Li2023BLIP2} and subsequently summarizing them into short video-level descriptions with an LLM. Panda-70M~\citep{chen2024panda} further scales supervision generation by employing eight different MLLMs to generate caption candidates. However, a set of generated captions is verified and selected by human annotators before training an automatic caption-selection model for final data generation.
Our framework is fully automated and generates multiple detailed and complementary textual views, providing richer supervision than short single-caption supervision.
\begin{figure*}[t]
    \centering
    \includegraphics[width=\linewidth]{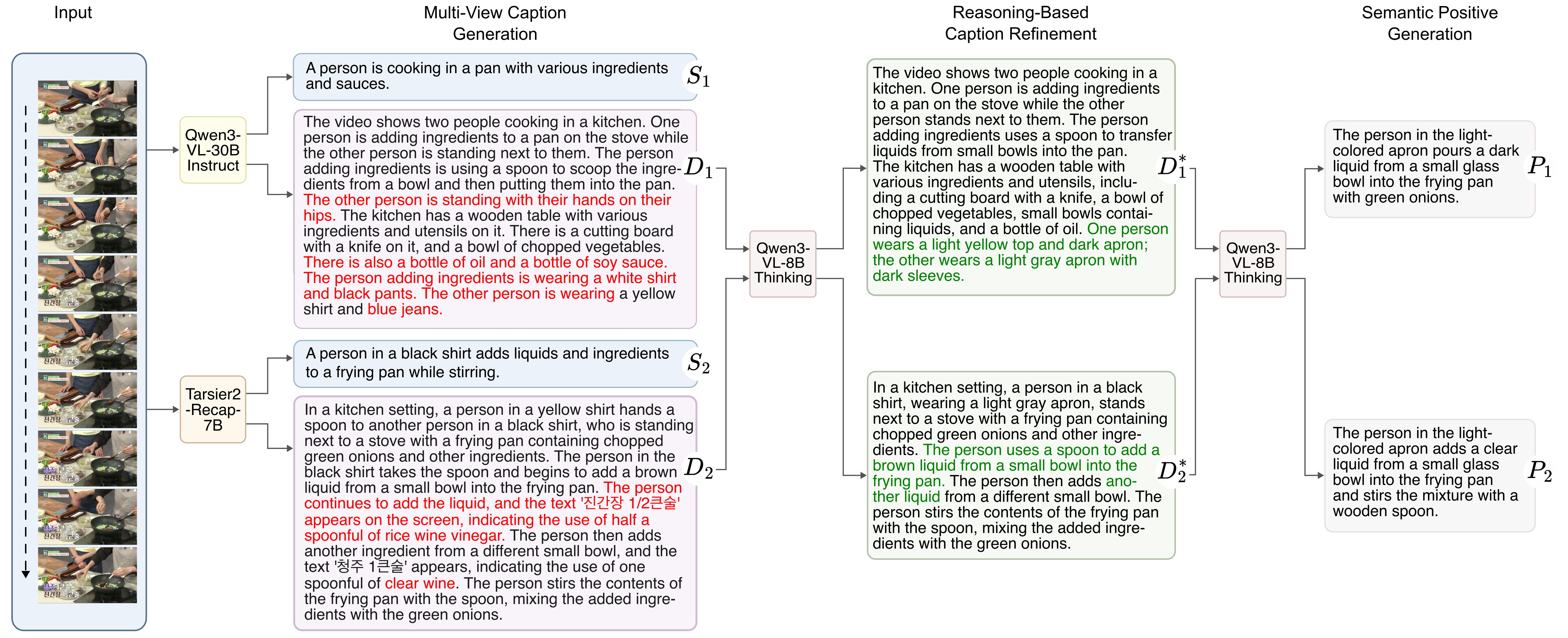}
    \vspace{-0.7cm}
    \caption{\textbf{Captioning pipeline.} Overview of our framework for improving supervision quality in video-text pretraining. Given an input video, we construct multiple complementary textual views, refine them through visual reasoning to improve grounding, and generate semantically consistent positives to enrich supervision.  \textcolor{red}{Red} highlights the potentially incorrect segments in the detailed caption, while \textcolor{green}{green} represents the corrected segments after refinement.
    The resulting supervision corpus is used to learn richer video-text representations.}
    \vspace{-0.5cm}
    \label{fig:method_overview} 
\end{figure*}
\section{Methodology}
\label{sec:methodology}

\vspace{-0.2cm}
\subsection{MLLM-Based Multi-View Supervision Generation}
\vspace{-0.2cm}
Large-scale video-text pretraining critically depends on the quality of video-text supervision. Existing web-scale datasets typically associate each video with a single caption, which often captures only a limited aspect of the visual content and overlooks rich spatiotemporal semantics. While recent multimodal large language models (MLLMs) can generate more informative descriptions, they may introduce hallucinations and unsupported details. We therefore design our supervision generation framework around three complementary objectives: semantic diversity, visual fidelity, and semantic coverage. Specifically, we generate multiple complementary captions, refine them through visual verification, and create additional semantic positive captions that capture different aspects of the same video. Together, these stages produce a supervision corpus that is diverse, visually faithful, and semantically comprehensive.

Starting from InternVid-10M-FLT~\citep{Wang2024InternVid}, we construct a large-scale multi-view supervision corpus through three stages: (i) multi-view caption generation, (ii) reasoning-based caption refinement, and (iii) semantic positive caption generation. Given an input collection of videos $\mathcal{V}=\{v_i\}_{i=1}^{N}$ with $N\approx10^7$, our goal is to generate multiple semantically complementary supervision signals for each video, as illustrated in Figure~\ref{fig:method_overview}. We denote summary captions by $S$, detailed captions by $D$, refined detailed captions by $D^{*}$, and semantic positive captions by $P$. Each stage increases either the diversity, fidelity, or semantic coverage of the supervision.

\vspace{-0.2cm}
\subsubsection{Multi-View Caption Generation.}
\vspace{-0.1cm}
Different MLLMs naturally generate diverse descriptions of the same video. We aim to leverage such captions from different MLLMs as complementary semantic views, providing richer supervision than any individual caption alone.
Let $f_m$ denote the captioning function of model $m$. Given a video $v_i$, each model generates both a concise summary and a detailed description,
\begin{equation}
\begin{aligned}
S_m &= f_m(v_i,q_s),\\
D_m &= f_m(v_i,q_d),
\end{aligned}
\end{equation}
where $q_s$ prompts the model to summarize the primary activity, while $q_d$ requests a richer description of visible objects, actions, interactions, and scene context.

We instantiate the caption generators using two pretrained video MLLMs, Tarsier2-Recap-7B~\citep{Yuan2025Tarsier2-arxiv} and Qwen3-VL-30B-Instruct~\citep{Bai2025Qwen3VL-arxiv}, producing two summary captions ($S_1,S_2$) and two detailed captions ($D_1,D_2$) for every video.

\vspace{-0.2cm}
\subsubsection{Reasoning-Based Caption Refinement.}
\vspace{-0.2cm}
Detailed captions provide richer supervision but frequently contain unsupported actions and artifacts such as subtitles, text overlays, user-interface elements, or incorrect information. We therefore introduce a reasoning-capable refinement stage that treats generated captions as noisy supervision candidates. Rather than generating a new description from scratch, the model acts as a verifier checking consistency between textual and visual evidence.

Given a detailed caption $D_m$ for $m\in\{1,2\}$, we employ a reasoning-capable MLLM $g$ based on Qwen3-VL-8B-Thinking~\citep{Bai2025Qwen3VL-arxiv}:
\begin{equation}
D_m^{*}=g(v_i,D_m),
\end{equation}
where the model jointly receives both the original video and the generated caption as input. We prompt the MLLM to preserve visually grounded content while removing unsupported details, potentially hallucinated actions, subtitles, watermarks, and non-visual inferences such as speech, intentions, or emotions. This stage, therefore, improves supervision fidelity while maintaining semantic richness. 
Although refinement substantially reduces incorrect information, some residual inaccuracies may still remain due to limitations of the underlying MLLM.
For each video $v_i$, this stage generates two refined detailed captions $D_1^{*}$ and $D_2^{*}$ from $D_1$ and $D_2$, respectively.

\vspace{-0.2cm}
\subsubsection{Semantic Positive Caption Generation.}
\vspace{-0.2cm}
Although refined detailed captions provide stronger visual grounding, a single description still captures only one semantic interpretation of a complex video. Multiple valid descriptions may exist depending on the event, interaction, object, or temporal phase being emphasized. We therefore generate additional semantic positive captions that shift the semantic focus while preserving visual correctness.

Unlike conventional paraphrasing, which primarily changes the language of the same description, semantic positive captions intentionally describe different yet visually grounded aspects of the same video. One may emphasize the manipulated object, another the interaction, and another the resulting state, while all remain semantically consistent with the underlying visual content. We randomly select one refined detailed caption $D_m^{*}$, where $m\in\{1,2\}$, and generate two semantic positive captions in one call:
\begin{equation}
(P_1,P_2)=h(v_i,D_m^{*}),
\end{equation}
where $h$ denotes Qwen3-VL-8B-Thinking~\citep{Bai2025Qwen3VL-arxiv}, conditioned jointly on the video and the selected refined detailed caption. Each semantic positive caption explicitly describes the actor, action, interacting object, and resulting outcome while emphasizing a different grounded aspect whenever possible.

The final caption set for video $v_i$ used for training is defined as:
\begin{equation}
\mathcal{C}(v_i) = \left\{S_1, S_2, D_1^{*},D_2^{*}, P_1,P_2,O \right\},
\end{equation}
Here $S_1$ and $S_2$ are the summary captions, $D_1^{*}$ and $D_2^{*}$ are the refined detailed captions, $P_1$ and $P_2$ are the semantic positive captions, and $O$ is the original InternVid-10M-FLT caption. The unrefined detailed captions $D_1$ and $D_2$ are intermediate outputs and are not included in the final training set. Detailed prompts, visual examples, and caption statistics are in the supplementary material.

\vspace{-0.2cm}
\subsection{Video-Text Pretraining}
\vspace{-0.2cm}
We adopt a two-stage video-text pretraining framework following recent video-text models~\citep{Li2023Unmasked}, while modifying the training pipeline to use the proposed multi-view supervision corpus.

\vspace{-0.2cm}
\paragraph{Stage I: Video Representation Initialization.}
We initialize the visual encoder using SMILE~\citep{thoker2025smile}, a masked video representation learning framework that jointly captures spatial semantics and motion dynamics through masked reconstruction objectives. Compared with conventional masked video pretraining, SMILE provides stronger motion-aware representations, which are particularly beneficial for downstream video-text alignment.

\vspace{-0.2cm}
\paragraph{Stage II: Video-Text Alignment.}
Starting from the pretrained visual encoder, we perform video-text pretraining using the supervision set $\mathcal{C}(v_i)$. We use two
\texttt{CLS} tokens in the text encoder: one for summary captions (including $O$) and one for refined detailed and semantic positive captions. During pretraining, we randomly sample for each video $v_i$ one caption for each \texttt{CLS} token:
\begin{equation}
\label{eq:sampling_summary_detailed}
c_s \sim \left\{S_1, S_2, O \right\}, \quad c_d \sim \left\{D_1^{*}, D_2^{*}, P_1, P_2\right\}.
\end{equation}
We encode the two captions separately, using the matching \texttt{CLS} token for each, and share the video representation between them. We train with three standard video-text objectives: video-text contrastive learning (VTC), video-text matching (VTM), and masked language modeling (MLM). For VTC and VTM, each caption is paired with the video using its corresponding \texttt{CLS} representation, and we apply MLM to both captions. For the summary view, the combined loss is
\begin{equation}
\mathcal{L}^{s} =
\mathcal{L}_{\mathrm{VTC}}^{s} +
\mathcal{L}_{\mathrm{VTM}}^{s} +
\mathcal{L}_{\mathrm{MLM}}^{s}.
\end{equation}
The detailed-view loss $\mathcal{L}^{d}$ is computed in the same way, using
the sampled refined detailed or semantic positive caption and its corresponding \texttt{CLS} token. We average the
summary-view and detailed-view losses to obtain the overall objective:
\begin{equation}
\mathcal{L} = \frac{1}{2}\left(\mathcal{L}^{s}+\mathcal{L}^{d}\right).
\end{equation}
During zero-shot text-to-video retrieval, we encode each text query twice, once with the summary \texttt{CLS} token and once with the detailed \texttt{CLS} token. We compute each text representation’s similarity to every candidate video and average the two similarity scores for each text-video pair. We then rank the videos by the averaged scores, following the UMT evaluation protocol.

Compared with conventional video-text pretraining pipelines that rely on a single caption per video, our framework exposes the model to multiple semantically consistent yet complementary textual views, enabling stronger alignment between visual content and language semantics.

\vspace{-0.2cm}
\section{Experiments}
\vspace{-0.2cm}
\label{sec:experiments}

\paragraph{Implementation Details:}
For Stage I, we use publicly available SMILE~\citep{thoker2025smile} checkpoints with ViT-B and ViT-L backbones pretrained on Kinetics-400 and Kinetics-700~\citep{Kay2017TheKinetics-arxiv}, respectively. For Stage II, we initialize the video encoder from Stage I and use pretrained BERT-base/BERT-large as the text encoder following~\citep{Li2023Unmasked}. We pretrain on 5M or 10M videos from InternVid-10M-FLT~\citep{Wang2024InternVid}, as indicated in each results table, using the supervision set $\mathcal{C}(v_i)$ defined in Section~\ref{sec:methodology}.
Each video is represented using 16 randomly sampled frames of resolution $224\times224$ with a video token masking ratio of $10\%$, while captions are truncated to 128 tokens. Training is conducted on 16 GPUs with batch size 128 per GPU for 20 epochs using AdamW ($lr=10^{-4}$, weight decay $0.02$). We adopt cosine decay with one warmup epoch and a final learning rate of $1\%$ of the initial value. The pretraining objective combines VTC, VTM, and MLM with equal weights, where VTM employs hard negative mining, VTC uses a temperature of $0.07$, and MLM applies a token masking ratio of $0.5$.
During pretraining, we sample one summary caption from $\{O,S_1,S_2\}$ and one refined detailed or semantic positive caption from $\{D_1^*,D_2^*,P_1,P_2\}$, and encode them using their corresponding summary-view and detailed-view \texttt{CLS} tokens. Unless otherwise specified, all main results use this dual-\texttt{CLS} configuration.

\vspace{-0.2cm}
\subsection{Standard Text-to-Video Retrieval}
\vspace{-0.2cm}
\paragraph{Datasets.}
We evaluate on five standard video-text retrieval benchmarks: MSR-VTT~\citep{xu2016msr}, DiDeMo~\citep{anne2017localizing}, ActivityNet~\citep{caba2015activitynet}, LSMDC~\citep{rohrbach2015dataset}, and MSVD~\citep{chen2011collecting}, following prior works.
MSR-VTT and MSVD focus on general semantic alignment over open-domain videos with diverse activities and scenes. DiDeMo and ActivityNet place greater emphasis on temporal reasoning and activity understanding, requiring the model to capture dynamically evolving content and long-range temporal dependencies. LSMDC presents a more challenging movie-based setting with complex narratives and richer contextual semantics. We strictly follow \citep{Li2023Unmasked} for both zero-shot and fine-tuning evaluation.  We report text-to-video retrieval performance using Recall (R@1). 

\vspace{-0.2cm}
\paragraph{Zero-Shot Results.}
Table~\ref{tab:zeroshot_retrieval_sources} presents zero-shot text-to-video retrieval results across five benchmarks. Under the 5M setting, our model outperforms prior approaches trained with comparable amounts of data. Compared with UMT-B, MVC-B improves by +8.6 on MSR-VTT, +25.5 on DiDeMo, and +30.3 on ActivityNet, while MVC-L similarly surpasses UMT-L by +27.2 on DiDeMo and +32.0 on ActivityNet. Compared with UMT pretrained on Panda-5M using captions generated by multiple cross-modality teachers, our approach achieves substantially stronger performance on MSR-VTT, DiDeMo, and MSVD using the same pretraining scale. These results suggest that richer supervision quality is more effective for learning video-text representations.

\begin{table*}[t]
\centering

\vspace{-0.5cm}
\caption{
\textbf{Zero-shot text-to-video retrieval.} We evaluate our method on MSR-VTT,  DiDeMo, ActivityNet, LSMDC, and MSVD using Recall@1. 
Img3M = CC3M,
Img15M = CC12M + SBU + COCO + VG. Within each pretraining-scale group, bold and underlined numbers represent the best and second-best models, respectively.
}

\scriptsize
\setlength{\tabcolsep}{2pt}

\resizebox{\textwidth}{!}{%
\begin{tabular}{l c c|c|c|c|c|c}
\toprule

\multirow{2}{*}{Method} 
& \multirow{2}{*}{\#Samples}
& \multirow{2}{*}{Source}
& \multicolumn{1}{c|}{MSR-VTT}
& \multicolumn{1}{c|}{DiDeMo}
& \multicolumn{1}{c|}{ActivityNet}
& \multicolumn{1}{c|}{LSMDC}
& \multicolumn{1}{c}{MSVD} \\

&
&
& R@1
& R@1
& R@1
& R@1
& R@1 \\

\midrule

Frozen~\citep{bain2021frozen} 
& 5M 
& WebVid-2M+Img3M
& 18.7 & 20.2 & -- & -- & -- \\

Singularity~\citep{lei2023revealing} 
& 5M 
& WebVid-2M+Img3M
& 28.4 & 36.9 & 30.8 & -- & -- \\

UMT-B~\citep{Li2023Unmasked}   
& 5M   
& WebVid-2M+Img3M
& 29.6 & 33.4 & 28.3 & 16.8 & 36.2 \\

STM-B~\citep{wu2025video}
& 5M 
& WebVid-2M+Img3M
& 29.8 & 34.4 & 30.6 & -- & 38.7 \\

ClusterSTM-B~\citep{zhuang2026cluster}
& 5M 
& WebVid-2M+Img3M
& 31.2 & 36.5 & 31.4 & -- & 40.3 \\

UMT-L~\citep{Li2023Unmasked}\footnotemark
& 5M 
& WebVid-2M+Img3M
& 33.3 & 34.0 & 31.9 & \underline{20.0} & \underline{44.4} \\

UMT-L(Panda-5M)~\citep{chen2024panda}
& 5M 
& Panda-5M
& 37.2 & 34.2 & -- & -- & 37.2 \\

\hdashline
Baseline-B
& 5M 
& InternVid-10M-FLT (original)
& 34.2 & 32.7 & 29.6 & 11.1 & 39.1 \\
MVC-B (Ours)
& 5M 
& InternVid-10M-FLT + MVC
& \underline{38.2} & \underline{58.9} & \underline{58.6} & 18.1 & 43.3 \\
MVC-L (Ours)
& 5M 
& InternVid-10M-FLT + MVC
& \textbf{40.6} & \textbf{61.2} & \textbf{63.9} & \textbf{21.0} & \textbf{48.4} \\

\midrule
VIOLET~\citep{fu2021violet} 
& 138M 
& HowTo100M
& 25.9 & 23.5 & -- & -- & -- \\

Singularity~\citep{lei2023revealing} 
& 17M 
& WebVid-2M+Img15M
& 34.0 & 37.1 & 30.6 & -- & -- \\

OmniVL~\citep{wang2022omnivl}
& 17M 
& WebVid-2M+Img15M
& 34.6 & 33.3 & -- & -- & -- \\

InternVideo~\citep{wang2022internvideo} 
& 646M 
& -
& 40.7 & 31.5 & 30.7 & 17.6 & 43.4 \\

VideoCoCa~\citep{yan2022videococa} 
& 4.8B 
& ALIGN+ALT200M
& 34.3 & -- & 34.5 & -- & -- \\

UMT-B ~\citep{Li2023Unmasked} 
& 17M 
& WebVid-2M+Img15M
& 35.5 & 41.9 & 33.8 & 18.1 & 41.4 \\

UMT-L~\citep{Li2023Unmasked} 
& 17M 
& WebVid-2M+Img15M
& \textbf{42.6} & {46.4} & {42.8} & \textbf{25.2} & \textbf{49.9} \\

CLIP~\citep{Radford2021Learning}
& 1B
& DataComp-1B
& 30.4 & 12.7 & 9.1 & 13.9 & 40.5 \\

ViCLIP-L~\citep{Wang2024InternVid}
& 410M
& CLIP-400M+InternVid-10M-FLT
& \underline{42.4} & 18.4 & 15.1 & 20.1 & \underline{49.1} \\

\hdashline
MVC-B (Ours)
& 10M 
& InternVid-10M-FLT + MVC
& 39.0 & \underline{61.0} & \underline{61.0} & 18.6 & 44.2 \\

MVC-L (Ours)
& 10M 
& InternVid-10M-FLT + MVC
& 41.1 & \textbf{62.0} & \textbf{66.9} & \underline{24.1} & 49.0 \\

\bottomrule
\end{tabular}
}

\label{tab:zeroshot_retrieval_sources}
\vspace{-0.3cm}

\end{table*}
\footnotetext{We report the corrected values from the official GitHub repository.}

More importantly, Baseline-B provides a controlled comparison that isolates the effect of our proposed supervision. Baseline-B uses the same backbone (with single-\texttt{CLS}), training framework, and 5M InternVid videos as MVC-B, but is trained with only the original captions $O$. Replacing the original supervision with MVC improves R@1 from 34.2 to 38.2 on MSR-VTT, 32.7 to 58.9 on DiDeMo, 29.6 to 58.6 on ActivityNet, 11.1 to 18.1 on LSMDC, and 39.1 to 43.3 on MSVD. The largest gains occur on DiDeMo (+26.2) and ActivityNet (+29.0), where retrieval requires matching videos to comparatively detailed descriptions of events and activities. This controlled comparison shows that the improvements arise substantially from enriching the textual supervision, rather than simply increasing the number of pretraining videos.

We further compare with methods pretrained on substantially larger corpora, ranging from 17M to 646M samples. Despite using only 10M videos, MVC-L remains competitive across standard retrieval benchmarks and achieves particularly strong performance on DiDeMo and ActivityNet. Compared with UMT-L pretrained on 17M videos, MVC-L improves R@1 by 15.6 points on DiDeMo and 24.1 points on ActivityNet while using fewer pretraining videos. Notably, MVC-L reaches 62.0 R@1 on DiDeMo and 66.9 on ActivityNet, compared with 31.5 and 30.7 for InternVideo trained on 646M samples. Together with the controlled Baseline-B comparison, these results suggest that richer multi-view supervision can achieve strong performance with substantially fewer pretraining videos.

\paragraph{Fine-Tuned Results.}
Table~\ref{tab:video_retrieval_r1_only} reports text-to-video retrieval after end-to-end fine-tuning. The improvements observed under zero-shot evaluation largely persist after task-specific adaptation. At the 5M scale, MVC-L substantially outperforms UMT-L on DiDeMo (+21.6), ActivityNet (+14.0), and LSMDC (+5.5), while also improving MSR-VTT (+2.2).

The controlled comparison with Baseline-B further shows that the benefits of our supervision persist after fine-tuning. With the same architecture and pretraining videos, MVC-B improves R@1 from 48.0 to 50.3 on MSR-VTT, 61.2 to 72.9 on DiDeMo, 55.7 to 64.4 on ActivityNet, 31.1 to 34.1 on LSMDC, and 45.0 to 46.0 on MSVD. Thus, the richer representations learned from multi-view supervision remain beneficial even after adaptation to individual downstream datasets.

The advantage also extends to the 10M setup. MVC-L achieves 58.1, 82.9, 75.2, and 45.6 R@1 on MSR-VTT, DiDeMo, ActivityNet, and LSMDC, respectively, outperforming UMT-L pretrained on 17M videos by 1.6, 16.3, 8.6, and 4.2 points. Similarly, compared with ViCLIP-L, which also incorporates InternVid-10M-FLT in its substantially larger pretraining corpus, MVC-L improves DiDeMo from 49.4 to 82.9 (+33.5) and ActivityNet from 49.8 to 75.2 (+25.4). These results further highlight the data efficiency of our approach, with particularly strong gains on benchmarks requiring retrieval from richer event and activity descriptions.

\begin{table*}[!t]
\centering

\vspace{-0.6cm}
\caption{
\textbf{Fine-Tuned Text-to-video retrieval.} We evaluate our method on MSR-VTT, DiDeMo, ActivityNet, LSMDC, and MSVD.  
Img3M  = CC3M
Img15M  = CC12M + SBU + COCO + VG
CC = Conceptual Captions,
VG = Visual Genome. Within each pretraining-scale group, bold and underlined numbers represent the best and second-best models, respectively.
}

\scriptsize
\setlength{\tabcolsep}{2pt}

\resizebox{\textwidth}{!}{%
\begin{tabular}{l c c|c|c|c|c|c}
\toprule

\multirow{2}{*}{Method} 
& \multirow{2}{*}{\#Samples}
& \multirow{2}{*}{Source}
& \multicolumn{1}{c|}{MSR-VTT}
& \multicolumn{1}{c|}{DiDeMo}
& \multicolumn{1}{c|}{ActivityNet}
& \multicolumn{1}{c|}{LSMDC}
& \multicolumn{1}{c}{MSVD} \\

&
&
& R@1
& R@1
& R@1
& R@1
& R@1 \\

\midrule
ClipBERT~\citep{lei2021less}
& 5.4M
& COCO+VG
& 22.0 & 20.4 & 21.3 & -- & -- \\

Frozen~\citep{bain2021frozen}
& 5M
& WebVid-2M+Img3M
& 31.0 & 34.6 & -- & 15.0 & 33.7 \\

UMT-B~\citep{Li2023Unmasked}
& 5M
& WebVid-2M+Img3M
& 46.3 & 54.8 & 52.1 & 30.3 & 47.4 \\

STM-B~\citep{wu2025video}
& 5M
& WebVid-2M+Img3M
& 48.5 & 56.9 & 53.6 & -- & -- \\

ClusterSTM-B~\citep{zhuang2026cluster}
& 5M
& WebVid-2M+Img3M
& 49.7 & 58.5 & 54.9 & -- & -- \\

UMT-L~\citep{Li2023Unmasked}
& 5M
& WebVid-2M+Img3M
& 53.3 & 59.7 & 58.1 & \underline{37.7} & 53.7 \\

UMT (Panda-5M)~\citep{chen2024panda}
& 5M
& Panda-5M
& \textbf{58.4} & 60.6 & -- & -- & \textbf{57.5} \\

\hdashline
Baseline-B
& 5M
& InternVid-10M-FLT (original)
& 48.0 & 61.2 & 55.7 & 31.1 & 45.0 \\

MVC-B (Ours)
& 5M
& InternVid-10M-FLT + MVC
& 50.3 & \underline{72.9} & \underline{64.4} & 34.1 & 46.0 \\

MVC-L (Ours)
& 5M
& InternVid-10M-FLT + MVC
& \underline{55.5} & \textbf{81.3} & \textbf{72.1} & \textbf{43.2} & \underline{53.8} \\

\midrule

VIOLET~\citep{fu2021violet}
& 138M
& HowTo100M
& 34.5 & 32.6 & -- & 16.1 & -- \\

All-in-one~\citep{wang2023all}
& 138M
& WebVid-2M+HowTo100M
& 37.9 & 32.7 & 22.4 & -- & -- \\

CLIP~\citep{Radford2021Learning,Wang2024InternVid}
& 400M
& CLIP-400M
& 38.2 & 32.2 & 26.1 & 22.5 & -- \\

LAVENDER~\citep{li2023lavender}
& 30M
& WebVid-2M+Vid13M+Img15M
& 40.7 & 53.4 & -- & 26.1 & 50.1 \\

Singularity~\citep{lei2023revealing}
& 17M
& WebVid-2M+Img15M
& 42.7 & 53.1 & 48.9 & -- & -- \\

VINDLU~\citep{Cheng2023VindLU}
& 25M
& WebVid-10M+Img15M
& 46.5 & 61.2 & 55.0 & -- & -- \\

HiTeA~\citep{ye2023hitea}
& 17M
& WebVid-2M+Img15M
& 46.8 & 56.5 & 49.7 & 28.7 & -- \\

OmniVL~\citep{wang2022omnivl}
& 17M
& WebVid-2M+Img15M
& 47.8 & 52.4 & -- & -- & -- \\

UMT-B~\citep{Li2023Unmasked}
& 17M
& WebVid-2M+Img15M
& 50.6 & 60.8 & 56.1 & 32.3 & 49.6 \\

UMT-B~\citep{Li2023Unmasked}
& 25M
& WebVid-2M+Img15M
& 51.0 & 61.6 & 58.3 & 32.7 & 50.8 \\

ViCLIP-L~\citep{Wang2024InternVid}
& 410M
& CLIP-400M+InternVid-10M-FLT
& 52.5 & 49.4 & 49.8 & 33.0 & 53.1 \\

CLIP-ViP~\citep{xue2022clip}
& 500M
& CLIP-400M+HD-VILA-100M
& 54.2 & 50.5 & 53.4 & 29.4 & -- \\

InternVideo~\citep{wang2022internvideo}
& 646M
& -
& {55.2} & 57.9 & 62.2 & 34.0 & -- \\

UMT-L~\citep{Li2023Unmasked}
& 17M
& WebVid-2M+Img15M
& \underline{56.5} & \underline{66.6} & \underline{66.6} & \underline{41.4} & \textbf{57.4} \\

\hdashline

MVC-B (Ours)
& 10M
& InternVid-10M-FLT + MVC
& 52.8 & \underline{74.8} & 65.2 & 35.2 & 50.1 \\

MVC-L (Ours)
& 10M
& InternVid-10M-FLT + MVC
& \textbf{58.1} & \textbf{82.9} & \textbf{75.2} & \textbf{45.6} & \underline{55.8} \\

\bottomrule
\end{tabular}
}

\label{tab:video_retrieval_r1_only}
\vspace{-0.3cm}

\end{table*}

\vspace{-0.5cm}
\subsection{Fine-Grained and Detailed Text-to-Video Retrieval}
\vspace{-0.2cm}
\paragraph{Datasets.}
Standard text-to-video retrieval benchmarks mainly evaluate alignment with summary-level descriptions, which may not adequately assess fine-grained visual understanding, temporal reasoning, or long-form semantics. Since our framework explicitly improves supervision quality through detailed, faithful, and diverse captions, we additionally evaluate zero-shot retrieval on CaReBench~\citep{xu2024carebench}, DREAM-1K~\citep{wang2024tarsier}, and Shot2Story20K~\citep{han2023shot2story20k}, which require richer text-to-video alignment. CaReBench evaluates spatial and temporal understanding through separate spatial and temporal retrieval settings. DREAM-1K focuses on fine-grained actions and event progressions, where we evaluate both detailed-description and event-based retrieval. Shot2Story20K consists of videos with multiple shots. We evaluate whole-clip retrieval and single-shot retrieval to assess long-form narrative semantics and local event understanding. We report text-to-video retrieval using Recall (R@1).

\vspace{-0.3cm}
\paragraph{Results.}
Table~\ref{tab:benchmark_comparison} presents zero-shot retrieval results on fine-grained and detailed benchmarks. The direct comparison between Baseline-B and MVC-B reveals particularly large gains in these more challenging retrieval settings. Using the same 5M videos, MVC improves R@1 by 38.2 points on CARE-S, 30.5 on CARE-T, 34.6 on DREAM-D, 10.6 on DREAM-E, 32.0 on S2S-W, and 29.1 on S2S-S. The consistent improvements across spatial, temporal, detailed-description, event-level, and multi-shot retrieval demonstrate the effectiveness of our framework for learning richer video-text correspondences. We separately isolate the contributions of multi-view supervision and granularity-aware text representations in Section~\ref{sec:ablations}.

The gains remain substantial compared with models trained on considerably more data. MVC-B pretrained on a 5M video subset outperforms UMT-L pretrained on 25M videos across all six settings, by 7.9 points on CARE-S, 13.0 on CARE-T, 11.0 on DREAM-D, 2.4 on DREAM-E, 18.8 on S2S-W, and 8.4 on S2S-S, despite using a smaller backbone and fewer videos. Scaling to MVC-L at the same 5M video scale further increases these margins to 9.4, 19.7, 12.6, 6.4, 20.4, and 11.3 points, respectively. These results show that the benefits of our framework become particularly pronounced when retrieval requires finer-grained or more descriptive video-text alignment.

This trend also holds against models pretrained on substantially larger corpora. Compared with ViCLIP-L, which additionally leverages CLIP-400M pretraining, MVC-L trained on 10M videos improves R@1 by 37.2 points on CARE-S, 32.9 on CARE-T, 40.9 on DREAM-D, 10.3 on DREAM-E, 41.3 on S2S-W, and 33.7 on S2S-S. MVC-L also consistently outperforms Long-CLIP-L across all six settings despite Long-CLIP's specialized long-text modeling. Together, these results support the complementary contributions of caption granularity, reasoning-based refinement, and semantic positive captions to the effectiveness of our supervision pipeline.

\begin{table*}[t]
\centering

\vspace{-0.5cm}
\caption{
\textbf{Fine-Grained and Detailed Zero-Shot Text-to-Video Retrieval.} We evaluate on CaReBench Spatial (CARE-S), CaReBench Temporal (CARE-T), DREAM-1K-Detailed (DREAM-D), DREAM-1K-Events (DREAM-E), and Shot2Story whole-clip (S2S-W) and single-shot (S2S-S) splits. Source labels use IV for InternVid and SG4V for ShareGPT4V. For prior methods, we use officially public checkpoints. Bold and underlined numbers mark the best and second-best results.
}

\small
\scriptsize
\setlength{\tabcolsep}{1pt}

\begin{tabular}{l|c|c|cccccc}
\toprule

Method 
& \#Samples
& Source
& CARE-S 
& CARE-T 
& DREAM-D 
& DREAM-E 
& S2S-W 
& S2S-S \\
&
&
& R@1
& R@1
& R@1
& R@1
& R@1
& R@1 \\

\midrule

CLIP-B/16~\citep{Radford2021Learning}
& 400M & CLIP-400M 
& 45.6 & 30.3 & 32.6 & 13.6 & 10.7 & 22.7 \\

CLIP-L/14~\citep{Radford2021Learning}
& 400M& CLIP-400M
& 49.1 & 33.5 & 44.3 & 14.6 & 65.8 & 45.4 \\

ViCLIP-B~\citep{Wang2024InternVid}
& 400M+ 10M & CLIP-400M + IV-10M-FLT
& 56.0 & 31.3 & 54.9 & 10.4 & 52.2 & 43.9 \\

ViCLIP-L~\citep{Wang2024InternVid}
& 400M + 10M& CLIP-400M + IV-10M-FLT
& 55.7 & 34.5 & 55.5 & 23.3 & 55.1 & 43.6 \\

Long-CLIP-L/14~\citep{zhang2024long}
& 400M& CLIP-400M + SG4V-1M
& 65.6 & 33.3 & 58.3 & 24.3 & 74.7 & 47.7 \\

UMT-B ~\citep{Li2023Unmasked}
& 5M& WebVid-2M + Img3M
& 64.2 & 36.5 & 70.0 & 18.0 & 64.6 & 45.8 \\

UMT-L~\citep{Li2023Unmasked}
& 5M  & WebVid-2M + Img3M
& 68.0 & 39.6 & 75.0 & 21.1 & 59.2 & 47.3 \\

UMT-B  ~\citep{Li2023Unmasked}
& 25M & WebVid-10M + Img15M
& 78.0 & 38.1 & 75.0 & 23.0 & 71.6 & 59.1 \\

UMT-L ~\citep{Li2023Unmasked}
& 25M
& WebVid-10M + Img15M
& 81.5 & 47.1 & 83.1 & 26.7 & 75.6 & 65.1 \\

\hdashline

Baseline-B & 5M & IV-10M-FLT (original)
& {51.2} & {29.6} & {59.5} & {18.5} & {62.4} & {44.4} \\

MVC-B (Ours)
& 5M & IV-10M-FLT + MVC
& {89.4} & {60.1} & {94.1} & {29.1} & {94.4} & {73.5} \\

MVC-B (Ours)
& 10M & IV-10M-FLT + MVC
& {90.6} & {60.9} & {94.5} & {30.1} & {95.2} & {75.0} \\

\hdashline
MVC-L (Ours)
& 5M& IV-10M-FLT + MVC
& \underline{90.9} & \underline{66.8} & \underline{95.7} & \underline{33.1} & \underline{96.0} & \underline{76.4} \\
MVC-L (Ours)
& 10M& IV-10M-FLT + MVC
& \textbf{92.9} & \textbf{67.4} & \textbf{96.4} & \textbf{33.6} & \textbf{96.4} & \textbf{77.3} \\

\bottomrule
\end{tabular}

\label{tab:benchmark_comparison}

\end{table*}

\begin{table*}[t]
\centering
\vspace{-0.5cm}
\caption{\textbf{Comparison of supervision recipes.}
We compare the original captions ($O$), our summary captions ($S$), our full generated supervision ($S+D^*+P$), and its combination with the original captions ($O+S+D^*+P$). We report text-to-video R@1 across six downstream benchmarks.}
\label{tab:supervision_recipes}
\setlength{\tabcolsep}{4pt}
\renewcommand{\arraystretch}{1.1}
\small
\begin{tabular}{lcccccc}
\toprule
Supervision
& MSVD & ANet & DiDeMo
& LSMDC & CARE-T & Shot2Story-S \\
\midrule
Original only ($O$)
& 35.4 & 25.9 & 29.3
& 11.8 & 27.1 & 42.2 \\

Ours: summary only ($S$)
& 37.0 & 46.2 & 46.0
& 15.1 & 48.0 & 63.2 \\

Ours: full ($S+D^*+P$)
& 38.5 & \textbf{50.1} & 52.4
& 16.7 & \textbf{54.1} & 67.4 \\

Original + ours ($O+S+D^*+P$)
& \textbf{38.8} & \textbf{50.1} & \textbf{53.0}
& \textbf{18.2} & \textbf{54.1} & \textbf{67.9} \\
\bottomrule
\end{tabular}
\vspace{-0.3cm}
\end{table*}

\begin{table*}[t]
\centering
\vspace{-0.5cm}
\caption{\textbf{Impact of caption complementarity, refinement, and semantic positive captions.} We compare summary captions ($S$), detailed captions ($D$), their combination, refined detailed captions ($D^*$), and semantic positive captions ($P$). Original captions ($O$) are excluded throughout.
We report text-to-video R@1.}
\label{tab:caption_components}
\setlength{\tabcolsep}{4pt}
\renewcommand{\arraystretch}{1.1}
\small
\begin{tabular}{lcccccc}
\toprule
Supervision
& MSVD & ANet & DiDeMo
& LSMDC & CARE-T & Shot2Story-S \\
\midrule
Summary only ($S$)
& 37.0 & 46.2 & 46.0
& 15.1 & 48.0 & 63.2 \\

Detailed only ($D$)
& 26.4 & 42.7 & 38.4
& 12.4 & 41.1 & 62.3 \\

Summary+detailed ($S+D$)
& 36.6 & 48.5 & 48.6
& 15.9 & 50.1 & {65.9} \\

Summary+refined detailed ($S+D^*$)
& {37.5} & 49.6 & 49.7
& {16.1} & 51.8 & 66.7 \\

Summary+refined+positive ($S+D^*+P$)
& \textbf{38.5} & \textbf{50.1} & \textbf{52.4}
& \textbf{16.7} & \textbf{54.1} & \textbf{67.4} \\
\bottomrule
\end{tabular}
\vspace{-0.3cm}
\end{table*}

\begin{table*}[ht]
\centering
\caption{\textbf{Impact of granularity-aware text representations.}
We compare a standard single-\texttt{CLS} representation with our dual-\texttt{CLS} design that separately models summary and detailed views. The detailed view includes refined detailed and semantic positive captions. Both use the same supervision ($O+S+D^*+P$) and training setup. We report text-to-video R@1.}
\label{tab:cls_ablation}
\setlength{\tabcolsep}{4pt}
\renewcommand{\arraystretch}{1.1}
\small
\begin{tabular}{lcccccc}
\toprule
Text Representation
& MSVD & ANet & DiDeMo
& LSMDC & CARE-T & Shot2Story-S \\
\midrule
Single \texttt{CLS}
& 38.8 & 50.1 & 53.0
& 18.2 & 54.1 & 67.9 \\

Dual \texttt{CLS} (Ours)
& \textbf{39.4} & \textbf{51.3} & \textbf{54.3}
& \textbf{18.7} & \textbf{56.4} & \textbf{68.5} \\
\bottomrule
\end{tabular}
\vspace{-0.3cm}
\end{table*}

\vspace{-0.2cm}
\subsection{Ablation Studies}
\vspace{-0.2cm}
\label{sec:ablations}

For computational efficiency, we conduct all ablations on a 1M-video subset of our pretraining corpus, use a video masking ratio of 50\%, while keeping the rest of the architecture, optimization, and training schedule fixed unless otherwise specified. We report zero-shot text-to-video R@1 on MSVD, ActivityNet, DiDeMo, LSMDC, CARE-T, and Shot2Story-S. The supervision ablations in Tables~\ref{tab:supervision_recipes} and~\ref{tab:caption_components} use a standard single-\texttt{CLS} representation, while Table~\ref{tab:cls_ablation} separately evaluates our dual-\texttt{CLS} design under identical supervision.

\vspace{-0.2cm}
\paragraph{Effect of Multi-View Supervision.}
Table~\ref{tab:supervision_recipes} studies the effect of progressively enriching the textual supervision. Replacing the original captions ($O$) with our summary captions ($S$) improves performance across all six benchmarks, with particularly large gains on ActivityNet (25.9$\rightarrow$46.2), DiDeMo (29.3$\rightarrow$46.0), CARE-T (27.1$\rightarrow$48.0), and Shot2Story-S (42.2$\rightarrow$63.2). This shows that summary captions generated from complementary MLLMs provide substantially more effective supervision than the original captions. Enriching these summary captions with refined detailed captions and semantic positive captions ($S+D^*+P$) further improves all six benchmarks, reaching 50.1 on ActivityNet, 52.4 on DiDeMo, 54.1 on CARE-T, and 67.4 on Shot2Story-S. Finally, retaining the original caption alongside our generated supervision ($O+S+D^*+P$) yields the best or equal-best performance across all benchmarks, including a notable improvement from 16.7 to 18.2 on LSMDC. These results suggest that the multi-view captions provide rich complementary supervision compared to the single-view captions across diverse downstream settings.

\vspace{-0.2cm}
\paragraph{Contribution of Supervision Components.}
Table~\ref{tab:caption_components} isolates the contributions of caption complementarity, refinement, and semantic positive captions. Detailed captions alone ($D$) perform worse than summary captions ($S$) across all benchmarks, indicating that detailed captions do not replace summary captions when used in isolation. Combining the two granularities ($S+D$), however, improves over $S$ alone on five of the six benchmarks, including ActivityNet (46.2$\rightarrow$48.5), DiDeMo (46.0$\rightarrow$48.6), and CARE-T (48.0$\rightarrow$50.1), demonstrating their complementary nature. Replacing the unrefined detailed captions with refined detailed captions ($S+D^*$) improves performance consistently across all six benchmarks, supporting the benefit of reasoning-based refinement. Adding semantic positive captions further improves every benchmark, with particularly clear gains on DiDeMo (49.7$\rightarrow$52.4) and CARE-T (51.8$\rightarrow$54.1). Together, these results validate the three objectives of our supervision pipeline: complementary caption granularities provide diverse semantic views, refinement improves visual fidelity, and semantic positive captions expand the semantic coverage of the supervision.

\vspace{-0.2cm}
\paragraph{Granularity-Aware Text Representations.}
Finally, Table~\ref{tab:cls_ablation} evaluates whether summary captions and refined detailed or semantic positive captions benefit from separate text representations. Both variants use identical $O+S+D^*+P$ supervision, isolating the effect of the representation design. The dual-\texttt{CLS} model consistently outperforms the standard single-\texttt{CLS} representation across all six benchmarks, improving ActivityNet from 50.1 to 51.3, DiDeMo from 53.0 to 54.3, and CARE-T from 54.1 to 56.4, with the largest gain of 2.3 points on CARE-T. These results show that explicitly modeling the summary and detailed views with separate representations provides an additional benefit beyond multi-view supervision alone, motivating the dual-\texttt{CLS} design used in our final model.

\vspace{-0.4cm}
\section{Conclusion}
\vspace{-0.2cm}
\label{sec:conclusion}
We revisit video-text pretraining from the perspective of supervision quality and propose a large-scale MLLM-based multi-view supervision framework for generating richer video-text annotations. By combining multi-view caption generation, reasoning-based refinement, and semantic positive generation, our approach improves supervision diversity, visual fidelity, and semantic coverage beyond conventional single-caption annotations. 
We further introduce a granularity-aware text representation that separately models short and detailed descriptions,
enabling the model to better exploit their complementary supervision. Extensive experiments across standard, fine-grained, and detailed video retrieval benchmarks demonstrate consistent gains in both zero-shot and fine-tuned settings, often surpassing approaches trained with larger datasets. 
Our results highlight the supervision quality and granularity as important dimensions of scalable video-text learning, demonstrating that richer textual supervision can significantly improve the data efficiency of video-text pretraining.

\section*{Acknowledgments}
The present research benefited from computational resources made available on Lucia, the Tier-1 supercomputer of the Walloon Region, infrastructure funded by the Walloon Region under the grant agreement n°1910247. We acknowledge LUMI-BE for awarding this project access to the LUMI supercomputer, owned by the EuroHPC Joint Undertaking, hosted by CSC (Finland) and the LUMI consortium through a LUMI-BE Regular Access call.
LUMI-BE is joint effort from BELSPO (federal), SPW Économie, Emploi, Recherche (Wallonia), Department of Economy, Science \& Innovation (Flanders) and Innoviris (Brussels). The research reported in this publication was supported by funding from King Abdullah University of Science and Technology (KAUST) - Center of Excellence for Generative AI, under award number 5940. For computing time, this research used Ibex managed by the Supercomputing Core Laboratory at King Abdullah University of Science \& Technology (KAUST) in Thuwal, Saudi Arabia.

\newpage
\appendix
\section{Appendix}
\label{sec:appendix}
\subsection{VQA Results}
\begin{table}[t]
\centering
\caption{\textbf{Video Question-Answering Results.}
We evaluate transfer to ActivityNet-QA, MSR-VTT-QA, and MSVD-QA. Our models show particularly strong performance on ActivityNet-QA while remaining competitive across the three benchmarks.}
\footnotesize
\setlength{\tabcolsep}{2pt}
\renewcommand{\arraystretch}{1.05}

\begin{tabular*}{.78\textwidth}{@{\extracolsep{\fill}}lcccc@{}}
\toprule
Method & \#Samples & ANet-QA & MSR-VTT-QA & MSVD-QA \\
\midrule

ClipBERT~\citep{lei2021less}
& 0.2M & -- & 37.4 & -- \\

ALPRO~\citep{li2022align}
& 5M & -- & 42.1 & 45.9 \\

JustAsk~\citep{yang2021just}
& 69M & 38.9 & 41.5 & 47.5 \\

VideoCLIP~\citep{xu2021videoclip}
& 136M & -- & -- & -- \\

All-in-one~\citep{wang2023all}
& 138M & -- & 44.3 & 47.9 \\

MERLOT~\citep{zellers2021merlot}
& 180M & 41.4 & 43.1 & -- \\

VIOLET~\citep{fu2021violet}
& 138M & -- & 43.9 & 47.9 \\

Singularity~\citep{lei2023revealing}
& 17M & 44.1 & 43.9 & -- \\

OmniVL~\citep{wang2022omnivl}
& 17M & -- & 44.1 & 51.0 \\

VINDLU~\citep{Cheng2023VindLU}
& 25M & 44.7 & 44.6 & -- \\

FrozenBiLM~\citep{yang2022zero}
& 400M & 43.2 & 47.0 & 54.8 \\

InternVideo~\citep{wang2022internvideo}
& 646M & -- & 47.1 & 55.5 \\

VideoCoCa~\citep{yan2022videococa}
& 4.8B & -- & 46.0 & 56.9 \\
\multirow{2}{*}{UMT-B~\citep{Li2023Unmasked}}
& 5M  & 43.5 & 44.3 & 49.1 \\
& 17M & 44.9 & 44.9 & 48.9 \\
\multirow{2}{*}{UMT-L~\citep{Li2023Unmasked}}
& 5M  & 45.1 & 45.5 & 51.3 \\
& 17M & 47.3 & 46.4 & 53.4 \\
\midrule
\multirow{2}{*}{MVC-B (Ours)}
& 5M & 47.3 & 45.1 & 49.5 \\
& 10M & 48.3 & 45.5 & 50.2 \\
\midrule
\multirow{2}{*}{MVC-L (Ours)}
& 5M & 49.1 & 46.0 & 51.5 \\
& 10M & 50.0 & 46.3 & 51.9 \\

\bottomrule
\end{tabular*}

\label{tab:vqa}
\end{table}

Table~\ref{tab:vqa} reports transfer performance on ActivityNet-QA (ANet-QA)~\citep{caba2015activitynet}, MSR-VTT-QA~\citep{xu2016msr}, and MSVD-QA~\citep{chen2011collecting}. Unlike retrieval, video question answering requires interpreting visual content in the context of a natural-language query, providing a complementary evaluation of the learned video-text representations.

For the 5M entries in Table~\ref{tab:vqa}, MVC improves over the corresponding UMT models on ActivityNet-QA. MVC-B improves from 43.5 to 47.3 (+3.8), while MVC-L improves from 45.1 to 49.1 (+4.0). The gains on MSR-VTT-QA are more modest, with MVC-L improving from 45.5 to 46.0, while performance on MSVD-QA remains comparable (51.3 vs. 51.5). Increasing the number of video clips from 5M to 10M further improves all three benchmarks, with MVC-L reaching 50.0 on ActivityNet-QA, 46.3 on MSR-VTT-QA, and 51.9 on MSVD-QA. Notably, MVC-L achieves the strongest ActivityNet-QA result in the table despite using only 10M video clips, exceeding UMT-L trained on 17M reported pretraining videos (50.0 vs. 47.3) and models trained on substantially larger corpora. Overall, these results indicate that the benefits of our richer textual supervision extend beyond retrieval to downstream video-text understanding, with particularly strong improvements on ActivityNet-QA.

\begin{figure*}[h]
    \centering
    \includegraphics[width=\linewidth]{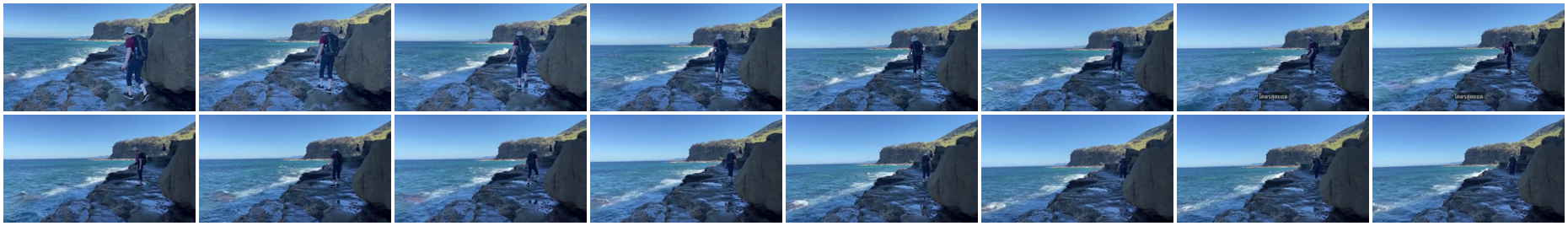}
    \caption{\textbf{Video frames.} Example of sub-sampled frames fed to the MLLMs for caption generation.}
    \label{fig:appendix_example_frames} 
\end{figure*}

\subsection{MLLM-Based Multi-View Supervision Generation}
We detail the prompts used to generate the different captions hereafter. For all types of captions, the video frames and the corresponding textual prompt are jointly fed to the MLLM as input. We use separate prompts for summary captions, detailed captions, refined detailed captions, and semantic positive captions, matching the stages described in the Methodology section.

\paragraph{Summary caption prompts.}
Summary captions are intended to provide a short semantic anchor for each video. We therefore ask each captioning MLLM to describe only the dominant activity in a single concise sentence.

\begin{promptbox}{Qwen3-VL-30B-Instruct}
Describe the main activity in this video, summarize the result in a single sentence of maximum 16 words.
\end{promptbox}

\begin{promptbox}{Tarsier2-Recap-7B}
Describe the main activity in this video in short.
\end{promptbox}

\paragraph{Detailed caption prompts.}
Detailed captions aim to expose the pretraining model to richer visual evidence than summary captions. The prompt therefore encourages the MLLM to describe the visible content in detail while keeping the output bounded.

\begin{promptbox}{Qwen3-VL-30B-Instruct}
Describe this video in detail, summarize the result in maximum of 128 words.
\end{promptbox}

\begin{promptbox}{Tarsier2-Recap-7B}
Describe the video in detail.
\end{promptbox}

\paragraph{Refined detailed caption prompts.}
Refined detailed captions are generated from each detailed caption independently. In each call, Qwen3-VL-8B-Thinking receives the video and one draft detailed caption, either from Qwen3-VL-30B-Instruct or Tarsier2-Recap-7B, and is asked to verify the caption against the visual content. The prompt emphasizes factual correction rather than free-form regeneration.

\begin{promptbox}{Qwen3-VL-8B-Thinking}
Analyze the visual content of this video focusing on the main action, objects, people, and environment. Be concrete and factual. Ignore text overlays, subtitles, watermarks, and UI elements. Do not infer audio, speech, emotions, or intentions.

Look at the following caption which has been generated by a video captioning model. Identify and correct any inaccuracies or hallucinations. Remove text overlays, watermarks, and UI elements mentioned in the caption. Here is the caption: \promptslot{$D_m$}.

Write the new caption, based on the one already provided, but corrected to accurately reflect the visual content of the video only. Remove any foreign characters or symbols unrelated to the core video content.
\end{promptbox}

\paragraph{Semantic positive caption prompts.}
Semantic positive captions are generated from a refined detailed caption and the video. Unlike generic paraphrases, these captions are constrained to describe specific visible events that remain semantically consistent with the same video. We ask Qwen3-VL-8B-Thinking to return two semantic positive captions by focusing on concrete actors, actions, objects, and outcomes. For each video, we randomly select one refined detailed caption $D_m^{*}$ with $m\in\{1,2\}$ and invoke the prompt once; the call returns $P_1$ and $P_2$.

\begin{promptbox}{Qwen3-VL-8B-Thinking}
Based on the visual content of the video and on the following detailed caption \promptslot{$D_m^{*}$}, provide two positive captions. Each positive caption must describe one specific, meaningful event that is clearly visible in the video, not a generic scene summary.

Each positive must explicitly include the actor, the key action, the object or person involved, and the visible outcome or state change. Make the two positives non-identical by focusing on different moments or phases of the event when possible, while remaining fully faithful to what is shown.

Use precise paraphrasing and action-level wording, and avoid vague text such as ``someone does something'' or purely static descriptions. Do not mention timestamps or frame numbers in the positives.
\end{promptbox}

\paragraph{Caption statistics.}
\Cref{tab:caption_stats} reports the mean and standard deviation of caption length in words for each intermediate and supervision view. The generated detailed captions are the longest and also show the largest spread, refinement reduces both their average length and variability, and the semantic positive captions remain concise while adding event-specific supervision.

\begin{table}[t]
\centering
\caption{\textbf{Caption statistics.} We report the average and standard deviation of caption length in words for each intermediate and supervision view.}
\label{tab:caption_stats}
\setlength{\tabcolsep}{4pt}
\small
\renewcommand{\arraystretch}{1.05}
\begin{tabular}{l c c}
\toprule
Caption type & Avg. words & Std. words \\
\midrule
Original caption & 9.4 & 2.6 \\
Qwen summary caption & 13.8 & 2.6 \\
Tarsier summary caption & 16.2 & 4.5 \\
Qwen detailed caption & 96.0 & 27.1 \\
Tarsier detailed caption & 80.1 & 25.0 \\
Qwen refined detailed caption & 60.6 & 19.5 \\
Tarsier refined detailed caption & 60.3 & 22.5 \\
Semantic positive caption & 25.0 & 7.6 \\
\bottomrule
\end{tabular}
\end{table}

These word-count differences reflect the roles of the caption types: summary captions provide concise anchors, detailed captions provide broader context, and semantic positive captions focus on specific visible events.

\paragraph{Effect of refined detailed captions.}
To complement the caption-length statistics, we analyze how refinement modifies detailed captions over a subset of 100,000 records. Given a detailed caption $D_m$ and its refined version $D_m^{*}$, we compute the Jaccard similarity over their content-word sets:
\[
J_{\mathrm{content}}(D_m,D_m^{*})
=
\frac{|C(D_m)\cap C(D_m^{*})|}
{|C(D_m)\cup C(D_m^{*})|},
\]
where $C(\cdot)$ denotes the set of content words. Higher values indicate greater lexical overlap. Since this metric ignores word order, frequency, and sentence structure, it measures lexical overlap rather than semantic correctness or claim preservation.

The average caption lengths observed on the audited subset are consistent with the aggregated statistics present in \Cref{tab:caption_stats}. Refinement reduces Qwen captions from 96.0 to 60.6 words and Tarsier captions from 80.1 to 60.3 words. No target length is imposed during refinement. Instead, as specified in the refinement prompt above, the model is instructed to remove text overlays, subtitles, watermarks, user-interface elements, unrelated symbols, hallucinated content, and non-visual inferences. The frequent shortening is therefore consistent with the intended removal of irrelevant or unsupported information. Qwen refinement yields a mean content-word Jaccard similarity of 0.405, with 0.47\% exact matches, whereas Tarsier refinement is more conservative, yielding a similarity of 0.564 and 2.58\% exact matches.

We further use Gemma-3-12B-IT as an independent text-only evaluator to assess the semantic relation between each detailed caption and its refined version. Gemma classifies 98.54\% of Qwen and 98.72\% of Tarsier transformations as semantically consistent or compatible, including omissions and changes in temporal focus. Specifically, 93.13\% of Qwen and 87.99\% of Tarsier refinements are classified as compatible with omissions, while only 0.79\% and 0.90\%, respectively, are classified as contradictory. These results indicate that refinement substantially compresses and rewrites the detailed captions while rarely introducing text-level semantic conflicts. Since Gemma observes only the captions, this analysis measures semantic consistency and transformation behavior rather than visual correctness.

\paragraph{Qualitative example.}
Figure~\ref{fig:appendix_example_frames} shows the video used in the example below. The original dataset caption is short and underspecified; each stage adds a complementary supervision signal with a different level of granularity or visual grounding.

\begin{supervisionbox}{Original dataset caption}
a person walks on rocks overlooking the ocean
\end{supervisionbox}

\begin{supervisionbox}{Summary captions}
\textbf{Qwen3-VL-30B-Instruct:} A person walks along a rocky shoreline, enjoying the ocean view and natural surroundings.

\textbf{Tarsier2-Recap-7B:} A person walks along a rocky path on a coastline.

\end{supervisionbox}

\begin{supervisionbox}{Detailed captions}
\textbf{Qwen3-VL-30B-Instruct:} A woman wearing a hat and backpack walks along a rocky shoreline. She holds a drone controller in her hand. The ocean is a deep blue and waves crash against the rocks. She walks away from the camera along the shore. She turns and walks back toward the camera. The sun is shining and the sky is clear. She walks toward the camera and then turns and walks away again. She stops and looks out at the ocean. The camera follows her as she walks along the shore.

\textbf{Tarsier2-Recap-7B:} A person wearing a hat, red shirt, and carrying a black backpack stands on a rocky coastline, looking at the ocean. The person begins to walk along the rocky path, moving away from the camera. The person continues walking, occasionally glancing back at the ocean. The person keeps walking along the rocky path, with the ocean and cliffs in the background.
\end{supervisionbox}

\begin{supervisionbox}{Refined detailed captions}
\textbf{Refined Qwen caption:} A person wearing a hat and backpack walks along a rocky shoreline. They hold a stick in their hand. The ocean is a deep blue with waves crashing against the rocks. The person walks away from the camera along the shore, stops, and continues walking away. The sun is shining and the sky is clear.

\textbf{Refined Tarsier caption:} A person wearing a hat, red shirt, and carrying a black backpack stands on a rocky coastline, looking at the ocean. The person begins to walk along the rocky path, moving away from the camera. The person continues walking, occasionally glancing back at the ocean. The person keeps walking along the rocky path, with the ocean and cliffs in the background.
\end{supervisionbox}

\begin{supervisionbox}{Semantic positive captions}
\textbf{Positive 1:} The person shifts weight from one foot to the other while stepping onto a wet rock, maintaining balance on the rocky path.

\textbf{Positive 2:} The person walks away from the camera, continuing along the rocky coastal trail toward the distant grassy cliff.
\end{supervisionbox}

\begin{figure*}[t]
    \centering
    \includegraphics[width=\linewidth]{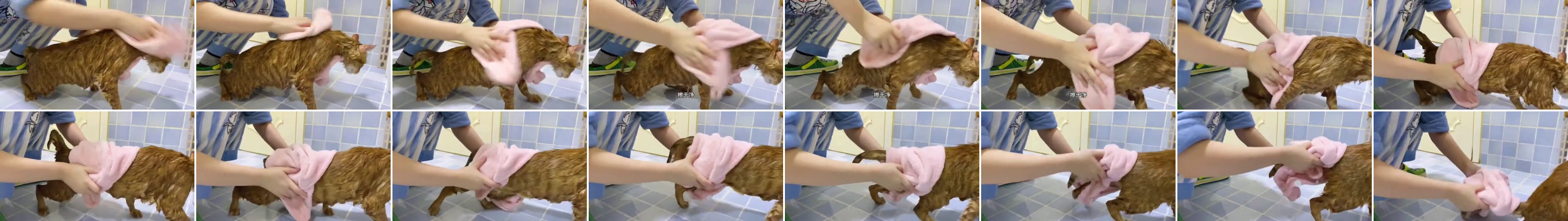}
    \caption{\textbf{Video frames.} Example of sub-sampled frames fed to the MLLMs for caption generation. This example shows the effect of refining the detailed captions to remove incorrect elements.}
    \label{fig:appendix_example_frames_refinement}
\end{figure*}

\paragraph{Refinement effectiveness example.}
\Cref{fig:appendix_example_frames_refinement} illustrates how refinement corrects object misidentification and removes unsupported actions in detailed captions. The original dataset caption and the Qwen3-VL-30B-Instruct captions incorrectly identify the animal as a dog, whereas the video shows a wet brown cat being dried with a pink towel. The detailed captions further introduce actions that are not supported by the visual content, such as lifting the animal's head or body. Refinement with Qwen3-VL-8B-Thinking corrects the animal category and removes these unsupported details in the detailed captions while preserving the visible drying activity. Incorrect or unsupported content in the input captions is highlighted in \textcolor{red}{red}, while corrected or visually grounded content in the refined detailed captions is highlighted in \textcolor{green}{green}.

\begin{supervisionbox}{Original dataset caption}
a small \textcolor{red}{dog} is being toweled in a small bathroom
\end{supervisionbox}

\begin{supervisionbox}{Summary captions}
\textbf{Qwen3-VL-30B-Instruct:} A person dries a wet
\textcolor{red}{dog} with a pink towel in a tiled bathroom.

\textbf{Tarsier2-Recap-7B:} A person dries a wet brown cat
with a pink towel in a tiled bathroom.
\end{supervisionbox}

\begin{supervisionbox}{Detailed captions}
\textbf{Qwen3-VL-30B-Instruct:} A person in a blue and white striped shirt and green shoes is drying a wet brown \textcolor{red}{dog} with a pink towel in a bathroom with blue tiles. The person wraps the towel around the \textcolor{red}{dog's} body and rubs it gently. The \textcolor{red}{dog} walks around while the person continues to dry it. \textcolor{red}{The person then lifts the dog's head and looks into its eyes.} The video ends with the person still drying the \textcolor{red}{dog}.

\textbf{Tarsier2-Recap-7B:} In a tiled bathroom, a person is seen drying a wet brown cat with a pink towel. The person starts by wiping the cat's back and head, moving the towel across the cat's body. The cat appears to be standing on a tiled floor, and the person continues to dry the cat, moving the towel from the back to the sides and front. The person then wraps the towel around the cat's body, \textcolor{red}{ensuring it is thoroughly dried. Finally, the person lifts the cat slightly, continuing to dry it with the towel.}
\end{supervisionbox}

\begin{supervisionbox}{Refined detailed captions}
\textbf{Refined Qwen caption:} A person in a blue and white striped shirt and green shoes is drying a \textcolor{green}{wet brown cat} with a pink towel in a bathroom with blue tiles. \textcolor{green}{The person rubs the towel over the cat's body and tail, moving it across the cat's fur.} The cat stands on the floor while the person continues to dry it. The video ends with the person still drying the cat.

\textbf{Refined Tarsier caption:} In a tiled room, a person is drying a wet brown cat with a pink towel. The person starts by wiping the cat's back, moving the towel across the cat's body. The person continues to dry the cat's sides, \textcolor{green}{then focuses on the tail area by moving the towel along the cat's back and sides. The person lifts the towel to dry the cat's tail.}
\end{supervisionbox}

\begin{supervisionbox}{Semantic positive captions}
\textbf{Positive 1:} A person wearing a blue and white striped shirt rubs a pink towel against the head of a brown tabby cat, with the cat remaining stationary on the blue-tiled floor while the towel makes contact with the cat's fur.

\textbf{Positive 2:} The same person shifts the pink towel to the back of the brown tabby cat, rubbing the towel over the cat's fur as the cat takes a step forward on the blue-tiled floor.
\end{supervisionbox}

\subsection{Datasets}
All datasets used in this work are publicly available and were used in accordance with their original licenses and terms of use. We will release the generated captions for the InternVid-10M-FLT dataset after acceptance.
\paragraph{InternVid.}
InternVid~\citep{Wang2024InternVid} is a large-scale video-text dataset designed for multimodal understanding and generation. It contains over 7M web videos spanning approximately 760K hours, resulting in 234M video clips paired with automatically generated textual descriptions. In this work, we use the publicly released InternVid-10M-FLT subset containing 10M filtered video clips with original captions for large-scale video-text pretraining. We refer to these clips as videos in the method and results sections.

\paragraph{MSR-VTT.}
MSR-VTT~\citep{xu2016msr} is a large-scale open-domain benchmark containing 10,000 web videos and approximately 200,000 captions covering diverse content such as sports, music, cooking, and daily activities. It is widely used for evaluating general video-text alignment. It contains around 7000 training and 1000 test samples.

\paragraph{DiDeMo.}
DiDeMo~\citep{anne2017localizing} consists of around 10,000 videos paired with multiple descriptions associated with temporally localized events. The benchmark primarily evaluates retrieval under temporal event understanding.  It contains around 8496 training and 1034 test samples.

\paragraph{ActivityNet Captions.}
ActivityNet Captions~\citep{caba2015activitynet} contains approximately 20,000 long videos with dense temporal annotations describing activities and events. The longer duration and multiple events per video make retrieval more challenging. It contains around 10K training and 5K test samples.

\paragraph{LSMDC.}
LSMDC~\citep{rohrbach2015dataset} contains around 118,000 movie clips paired with text derived from scripts and audio descriptions. The benchmark includes complex scenes and narrative content from movies.

\paragraph{MSVD.}
MSVD~\citep{chen2011collecting} contains roughly 2,000 short videos with around 80,000 human-written captions. Despite its smaller size, it remains a commonly used benchmark for video-text evaluation.

\paragraph{CaReBench.}
CaReBench~\citep{xu2024carebench} is a fine-grained retrieval benchmark providing separate spatial and temporal annotations, enabling independent evaluation of appearance-focused and temporal retrieval. The benchmark contains 1,000 test samples.

\paragraph{DREAM-1K.}
DREAM-1K~\citep{wang2024tarsier} contains 1,000 test videos paired with detailed descriptions of actions and event sequences. We evaluate both detailed-description retrieval and event-based retrieval settings.

\paragraph{Shot2Story20K.}
Shot2Story20K~\citep{han2023shot2story20k} focuses on multi-shot video understanding using shot-level captions and video-level summaries. We use the 2,000-sample test split and evaluate both whole-clip and single-shot retrieval settings.

\subsection{Additional Ablations}
\paragraph{Effect of Masking Ratio.}
Table~\ref{tab:masking_ablation} studies the effect of the video masking ratio while keeping the caption supervision fixed. Reducing the masking ratio from 50\% to 10\% improves performance on MSVD (38.8$\rightarrow$39.7), ActivityNet (50.1$\rightarrow$52.5), DiDeMo (53.0$\rightarrow$54.3), and CARE-T (54.1$\rightarrow$55.9), while performance on Shot2Story-S remains similar. In contrast, LSMDC favors the higher masking ratio, decreasing from 18.2 to 15.7 at 10\%. Overall, a lower masking ratio is beneficial on most benchmarks, suggesting that retaining more visual tokens is advantageous when learning from richer textual supervision.

\begin{table*}[h]
\centering
\caption{\textbf{Effect of Masking Ratio.}
We compare different video masking ratios while retaining the same
caption supervision ($O+S+D^*+P$). We report text-to-video R@1;
bold indicates the best result in each column.}
\label{tab:masking_ablation}
\setlength{\tabcolsep}{6pt}
\renewcommand{\arraystretch}{1.1}
\small
\begin{tabular}{lcccccc}
\toprule
Masking ratio
& MSVD & ANet & DiDeMo & LSMDC & CARE-T & Shot2Story-S \\
\midrule
50\%
& 38.8 & 50.1 & 53.0 & \textbf{18.2} & 54.1 & 67.9 \\

25\%
& 39.5 & 51.9 & 53.0 & 15.8 & 53.3 & \textbf{68.5} \\

10\%
& \textbf{39.7} & \textbf{52.5} & \textbf{54.3}
& 15.7 & \textbf{55.9} & 68.3 \\
\bottomrule
\end{tabular}
\end{table*}

\begin{table*}[h]
\centering
\caption{\textbf{Effect of training duration.}
We evaluate different pretraining durations while keeping the model,
supervision, and training configuration fixed. We report text-to-video
R@1 across six downstream benchmarks.}
\label{tab:training_schedule}
\setlength{\tabcolsep}{6pt}
\renewcommand{\arraystretch}{1.1}
\small

\begin{tabular}{lcccccc}
\toprule
Epochs
& MSVD & ANet & DiDeMo & LSMDC & CARE-T & Shot2Story-S \\
\midrule
5
& 31.6 & 40.8 & 42.6
& 12.9 & 42.9 & 59.7 \\

10
& 36.2 & 46.5 & 48.7
& 15.8 & 50.3 & 64.9 \\

15
& 37.9 & 48.9 & 51.5
& 17.3 & 52.6 & 66.7 \\

20
& \textbf{38.8} & \textbf{50.1} & \textbf{53.0}
& \textbf{18.2} & \textbf{54.1} & \textbf{67.9} \\
\bottomrule
\end{tabular}
\end{table*}

\paragraph{Effect of Training Schedule.} Table~\ref{tab:training_schedule} evaluates different training durations. Performance improves through 20 epochs, with smaller gains from 15 to 20 epochs than at earlier intervals. This differs from many previous video-text pretraining settings that commonly adopt shorter schedules, indicating that richer multi-view supervision can continue providing useful learning signals over longer optimization periods.

\subsection{Fine-Tuning Hyperparameters}
Table~\ref{tab:finetune_settings} shows the fine-tuning settings for text-to-video retrieval and VQA. All reported results correspond to a single training run following standard evaluation protocols used in prior works.
\begin{table}[t]
\centering

\caption{\textbf{Fine-Tuning Hyperparameters.} Settings used for retrieval and VQA experiments across datasets.}

\footnotesize
\setlength{\tabcolsep}{4pt}
\renewcommand{\arraystretch}{1.05}

\begin{tabular*}{.8\textwidth}{@{\extracolsep{\fill}}lcccc}
\toprule
Task & Dataset & LR & Epochs & DropPath \\
\midrule

\multirow{5}{*}{Retrieval}
& MSR-VTT     & 2e-5 (B), 1e-5 (L)              & 10(B),7(L)  & 0.2(B),0.3(L) \\
& DiDeMo     & 2e-5(B), 2e-5(L)         & 12(B),5(L)  & 0.1(B),0.3(L) \\
& ActivityNet& 4e-5(B), 1e-5 (L)               & 20(B/L)     & 0.1(B),0.2(L) \\
& LSMDC      & 2e-5 (B), 1e-5 (L)               & 10(B),8(L)  & 0.1(B),0.2(L) \\
& MSVD       & 2e-5 (B), 1e-5 (L)               & 10(B/L)     & 0.2(B),0.3(L) \\

\midrule

\multirow{3}{*}{VQA}
& ActivityNet-QA & 4e-5(B),1e-5(L) & 12(B),10(L) & 0.2(B),0.3(L) \\
& MSR-VTT-QA     & 2e-5(B),1e-5(L) & 8(B/L)      & 0.2(B),0.4(L) \\
& MSVD-QA        & 2e-5(B),1e-5(L) & 15(B),6(L)  & 0.2(B),0.4(L) \\

\midrule

Optimizer & \multicolumn{4}{c}{AdamW} \\
Momentum & \multicolumn{4}{c}{$\beta_1,\beta_2=0.9,0.999$} \\
Weight decay & \multicolumn{4}{c}{0.02} \\
LR schedule & \multicolumn{4}{c}{Cosine decay} \\
Batch size & \multicolumn{4}{c}{256} \\
Warmup epochs & \multicolumn{4}{c}{1} \\
Input frames & \multicolumn{4}{c}{12} \\
Augmentation & \multicolumn{4}{c}{MultiScaleCrop [0.5,1], Flip} \\

\bottomrule
\end{tabular*}

\label{tab:finetune_settings}
\end{table}

\end{document}